\documentclass[11pt]{article}
\usepackage[margin=1in]{geometry}
\usepackage{amsmath,amssymb}
\usepackage[numbers]{natbib}
\usepackage[colorlinks=true]{hyperref}
\usepackage{graphicx}
\title{Unifying ICL, SFT, KL-Regularized RL Through a Bayesian Lens}
\author{Junxin Fan\\
Fudan University\\
\texttt{22307130041@m.fudan.edu.cn}}
\date{\today{}}

\begin{document}
\maketitle
\begin{abstract}
Supervised fine-tuning (SFT), few-shot in-context learning (ICL), KL-regularized
RLHF/RLVR, and on-policy distillation are usually treated as distinct post-training
paradigms. We develop a unified Bayesian perspective in which each is an instance of
a two-step template: construct a (generalized) Bayes or Gibbs posterior from a
reference model and a utility signal (log-likelihood, reward, or advantage), then
approximate it by a forward-KL projection onto a parametric family, either
in-weights (SFT/RL) or in-context (ICL). This yields a single chain of equivalences:
few-shot ICL is an amortized projection onto the Bayes posterior predictive, and
reward-weighted SFT, reward-weighted ICL, and advantage-weighted SFT are forward-KL
projections of reward-induced Gibbs posteriors. The framework explains why
supervised warm-up is practically unavoidable for importance-weighted projections,
and interprets R1/o1-style reasoning models as combining test-time Bayesian search
with training-time amortization. Matched-budget experiments on Qwen3 models
corroborate the picture: operators that share their learning-signal granularity
produce nearly identical updates when support is good and diverge when it degrades,
and reward-weighted projection performs on par with standard baselines.
\end{abstract}


\section*{Part I: Few-shot ICL $\approx$ SFT (Bayesian View)}

\subsection*{1.1 Supervised Fine-Tuning (SFT) as Maximum Likelihood}

Given a dataset $\mathcal{D} = \{(x_i,y_i)\}_{i=1}^N$, standard supervised fine-tuning maximizes the conditional log-likelihood:
\begin{equation}
  \theta_{\mathrm{SFT}}
  \;=\;
  \arg\max_\theta \sum_{i=1}^N \log p_\theta(y_i \mid x_i).
\end{equation}
Equivalently, define the empirical data distribution
\begin{equation}
  \hat p_{\mathrm{data}}(x,y)
  \;=\;
  \frac{1}{N}\sum_{i=1}^N \delta_{(x_i,y_i)}(x,y),
\end{equation}
so that the SFT objective can be written as
\begin{equation}
  \theta_{\mathrm{SFT}}
  \;=\;
  \arg\min_\theta
  \mathbb{E}_{(x,y)\sim \hat p_{\mathrm{data}}}
  \big[-\log p_\theta(y\mid x)\big].
\end{equation}

\paragraph{Bayesian / KL interpretation.}
Assume the data are generated from an unknown teacher conditional distribution $q_{\mathrm{true}}(y\mid x)$, and that $\hat p_{\mathrm{data}}$ is an empirical approximation to the joint $q_{\mathrm{true}}(x,y) = q_{\mathrm{true}}(x)\,q_{\mathrm{true}}(y\mid x)$. Then
\begin{align}
  \mathbb{E}_{(x,y)\sim q_{\mathrm{true}}}\big[-\log p_\theta(y\mid x)\big]
  &= \mathbb{E}_{x\sim q_{\mathrm{true}}(x)}
     \mathbb{E}_{y\sim q_{\mathrm{true}}(\cdot\mid x)}\big[-\log p_\theta(y\mid x)\big] \\
  &= \mathbb{E}_{x\sim q_{\mathrm{true}}(x)}
     \Big[
       H\big(q_{\mathrm{true}}(\cdot\mid x)\big)
       + \mathrm{KL}\big(q_{\mathrm{true}}(\cdot\mid x)\,\|\,p_\theta(\cdot\mid x)\big)
     \Big],
\end{align}
Here we expand the inner expectation using the standard cross-entropy
decomposition $H(q,p)=H(q)+\mathrm{KL}(q\|p)$ applied to
$q_{\mathrm{true}}(\cdot\mid x)$ and $p_\theta(\cdot\mid x)$,
where $H(\cdot)$ is conditional entropy. The first term is independent of $\theta$, so minimizing expected negative log-likelihood is equivalent to
\begin{equation}
  \theta^\ast
  \;=\;
  \arg\min_\theta
    \mathbb{E}_{x\sim q_{\mathrm{true}}(x)}
    \mathrm{KL}\big(q_{\mathrm{true}}(\cdot\mid x)\,\|\,p_\theta(\cdot\mid x)\big).
\end{equation}
In words, SFT learns $p_\theta$ as the forward KL (maximum-likelihood) projection of the teacher $q_{\mathrm{true}}$ onto the model family $\{p_\theta\}$.\footnote{Here ``projection'' refers to minimizing the forward KL $\mathrm{KL}(q_{\mathrm{true}}\|p_\theta)$, also known as MLE or the m-projection in information geometry.}

\subsection*{1.2 Hierarchical Bayesian model for tasks and contexts}

We now introduce a hierarchical Bayesian model for tasks, following the ``ICL is Bayes''  framework of \citet{WakayamaBayesICL}.

\paragraph{Task-level generative model.}
Let $\mathcal{F}$ be a space of functions $f : \mathcal{X} \to \Delta(\mathcal{Y})$. A random task is sampled as
\begin{equation}
  f \sim P(f),
\end{equation}
and conditioned on $f$, data pairs $(x,y)$ are generated i.i.d. by
\begin{equation}
  x \sim P_X,
  \qquad
  y \sim f(\cdot\mid x).
\end{equation}
For a given task $f$, a few-shot context (prompt) and a query point are generated as
\begin{equation}
  D_k = \{(x_i,y_i)\}_{i=1}^k,
  \qquad
  (x_{k+1},y_{k+1}) \sim P_X \times f(\cdot\mid x_{k+1}).
\end{equation}

\paragraph{Bayes posterior over tasks and posterior predictive.}
Given context $D_k$, the posterior over tasks is
\begin{equation}
  P(f \mid D_k)
  \;\propto\;
  P(f)\,\prod_{i=1}^k f(y_i\mid x_i),
\end{equation}
and the corresponding Bayes posterior predictive at a new query $x_{k+1}$ is
\begin{equation}
  P_{\mathrm{Bayes}}(y\mid x_{k+1}, D_k)
  \;:=\;
  \mathbb{E}_{f\sim P(f\mid D_k)}\big[f(y\mid x_{k+1})\big].
\end{equation}

\paragraph{Bayes-optimal in-context predictor.}
Consider any in-context prediction rule $M$ that maps $(D_k,x_{k+1})$ to a predictive distribution
\(
  M(D_k,x_{k+1}) \in \Delta(\mathcal{Y}).
\)
Under log-loss, define the ICL risk
\begin{equation}
  R(M)
  \;:=\;
  \mathbb{E}\big[-\log M(D_k,x_{k+1})(y_{k+1})\big],
\end{equation}
where the expectation is over the hierarchical model above (sampling $f$, then $D_k$ and $(x_{k+1},y_{k+1})$).

\textbf{Theorem 1 (Bayes-optimal in-context predictor; after \citet{WakayamaBayesICL}).}
Define the Bayes posterior predictive
\begin{equation}
  M_{\mathrm{Bayes}}(D_k,x_{k+1})
  := P_{\mathrm{Bayes}}(\cdot\mid x_{k+1}, D_k).
\end{equation}
Then:
\begin{itemize}
  \item[(i)] $M_{\mathrm{Bayes}}$ is the unique minimizer of $R(M)$ over all measurable predictors $M$.
  \item[(ii)] For any $M$, we define its Bayes Gap as
  \begin{equation}
    \mathrm{BayesGap}(M)
    \;:=\;
    R(M) - R(M_{\mathrm{Bayes}}).
  \end{equation}
  In particular, the excess risk satisfies
  \begin{equation}
    R(M) - R(M_{\mathrm{Bayes}})
    \;=\;
    \mathrm{BayesGap}(M).
  \end{equation}
  In the squared-loss setting of \citet{WakayamaBayesICL}, this quantity coincides with their
  Bayes Gap term $R_{\mathrm{BG}}(M)$ in the decomposition $R(M)=R_{\mathrm{BG}}(M)+R_{\mathrm{PV}}$,
  and $R(M_{\mathrm{Bayes}})$ corresponds to their Posterior Variance component $R_{\mathrm{PV}}$.

\end{itemize}

The full non-asymptotic bounds for the Bayes Gap and Posterior Variance are given in \citet{WakayamaBayesICL}; here we only use that $M_{\mathrm{Bayes}}$ is the normative target of in-context prediction. This hierarchical model can be justified from a within-task exchangeability assumption via a de Finetti–type representation; see Appendix A.4.

\paragraph{Remark (Order of demonstrations).}
Real transformers process ordered sequences with positional embeddings, so the order of
few-shot demonstrations in the prompt can affect predictions. Our exchangeability
assumption applies at the level of the \emph{data-generating process}: within a task,
the examples $(x_i,y_i)$ are assumed to form an exchangeable sample from an underlying
task distribution, so the Bayes posterior $P(f\mid D_k)$ and posterior predictive
$P_{\text{Bayes}}(y\mid x_{k+1},D_k)$ depend only on the multiset $\{(x_i,y_i)\}$, not on
their order. Part~I analyzes this ideal, order-symmetric Bayes predictor; practical
in-context learners may introduce position-dependent biases, which we view as algorithmic
deviations from this ideal rather than violations of the assumption.

\subsection*{1.3 In-context learning as amortized Bayesian inference}

Let $M_\theta(y\mid x,D_k)$ denote the conditional distribution implemented by a pretrained Transformer with parameters $\theta$. Pretraining across many tasks and contexts can be abstracted as minimizing the expected log-loss
\begin{equation}
  R(\theta)
  \;:=\;
  \mathbb{E}\big[-\log M_\theta(D_k,x_{k+1})(y_{k+1})\big],
\end{equation}
where the expectation is over the same hierarchical process as in Section~1.2.

By Theorem~1, the risk-minimizing population predictor is $M_{\mathrm{Bayes}}$. Thus, in the limit of infinite model capacity and pretraining data, any minimizer $M_{\theta^\ast}$ of $R(\theta)$ must satisfy
\begin{equation}
  M_{\theta^\ast}(D_k,x_{k+1})
  \;\approx\;
  M_{\mathrm{Bayes}}(D_k,x_{k+1})
  \;=\;
  P_{\mathrm{Bayes}}(\cdot\mid x_{k+1}, D_k).
\end{equation}
Formally, under the assumption in Sec.~1.5 that the model class $\{M_\theta\}$ is rich enough to approximate $M_{\mathrm{Bayes}}$, any global minimizer (or sequence of asymptotic minimizers) of $R(\theta)$ can be chosen to be arbitrarily close to $M_{\mathrm{Bayes}}$ in expected log-loss.

This viewpoint matches recent constructions that explicitly train Transformers as amortized Bayesian inference procedures over latent variables and datasets (e.g., \citep{ReuterBayesTransformer}), where a single forward pass maps each dataset $D$ to an approximate posterior.

\paragraph{Interpretation.}
In this Bayesian picture, few-shot ICL is best viewed as learning the mapping
\begin{equation}
  (D_k,x_{k+1})
  \;\longmapsto\;
  P_{\mathrm{Bayes}}(\cdot\mid x_{k+1}, D_k)
\end{equation}
in an amortized way. The Transformer does not explicitly perform gradient descent in parameter space at test time; instead, it approximates the Bayes posterior predictive directly in its forward pass. While our exposition centers on the Bayesian view, a complementary first-order / meta-learning interpretation (ICL $\approx$ one GD step) has been
studied extensively in prior work \citep{DaiICLMetaOpt,AkyurekICLLinear,LiTransAsAlgo}, which shows that Transformers can implement
gradient-based updates in their forward passes. We do not re-derive those results here, as our
goal is to foreground the Bayesian formulation.

\subsection*{1.4 From Bayes posterior to SFT: forward-KL projection over outputs}

We now connect the Bayes posterior predictive to supervised fine-tuning.

Fix a context $D_k$ and query $x_{k+1}$, and write
\begin{equation}
  q_{\mathrm{Bayes}}(y\mid x_{k+1}, D_k)
  \;:=\;
  P_{\mathrm{Bayes}}(y\mid x_{k+1}, D_k)
\end{equation}
for the ideal target distribution over outputs. Consider any parametric conditional model $p_\theta(y\mid x)$ (suppressing $D_k$ in notation). For this fixed pair $(x_{k+1},D_k)$, the best approximation to $q_{\mathrm{Bayes}}$ within $\{p_\theta\}$ under log-loss is
\begin{equation}
  \theta^\ast(x_{k+1},D_k)
  \;=\;
  \arg\min_\theta
  \mathrm{KL}\big(q_{\mathrm{Bayes}}(\cdot\mid x_{k+1},D_k)\,\|\,p_\theta(\cdot\mid x_{k+1})\big),
\end{equation}
or equivalently
\begin{equation}
  \theta^\ast(x_{k+1},D_k)
  \;=\;
  \arg\max_\theta
  \sum_y q_{\mathrm{Bayes}}(y\mid x_{k+1},D_k)\,\log p_\theta(y\mid x_{k+1}).
\end{equation}

If we had direct samples $y \sim q_{\mathrm{Bayes}}(\cdot\mid x_{k+1},D_k)$ for many $(x_{k+1},D_k)$, then performing SFT on those samples would be exactly the forward KL projection of the Bayes posterior predictive onto the family $\{p_\theta\}$. Aggregating over contexts and queries yields the population objective
\begin{equation}
  \min_\theta
  \mathbb{E}_{(D_k,x_{k+1})}
  \mathrm{KL}\big(q_{\mathrm{Bayes}}(\cdot\mid x_{k+1},D_k)\,\|\,p_\theta(\cdot\mid x_{k+1})\big),
\end{equation}
and the standard SFT objective is its empirical approximation.

\paragraph{Bayesian reading of ``ICL $\approx$ SFT''.}
Putting Sections~1.2-1.4 together:
\begin{itemize}
  \item The ideal in-context predictor is the Bayes posterior predictive $q_{\mathrm{Bayes}}(\cdot\mid x_{k+1},D_k)$.
  \item A sufficiently expressive Transformer minimizing ICL risk implements an amortized approximation $M_\theta(\cdot\mid x_{k+1},D_k)$ to $q_{\mathrm{Bayes}}$.
  \item Supervised fine-tuning (SFT) learns $p_\theta$ as the forward KL projection of $q_{\mathrm{Bayes}}$ (or of an empirical approximation) onto the parametric family.
\end{itemize}
Thus, in the Bayesian view,
\begin{equation}
  \text{few-shot ICL}
  \;\approx\;
  \text{Bayes posterior predictive}
  \;\approx\;
  \text{KL-projected SFT solution},
\end{equation}
and "ICL $\approx$ SFT" is understood as both being different approximations to the same underlying Bayesian predictor.

This distribution-level identification is the backbone for Part~II, where KL-regularized RL (RLHF/RLVR) is shown to define a generalized Gibbs posterior over outputs, and reward-weighted SFT is exactly its KL projection onto $\{p_\theta\}$.

\subsection*{1.5 Assumptions and scope}

\paragraph{Regularity for Part~I (Bayesian ICL and SFT).}
We assume:
(i) a well-defined hierarchical prior $P(f)$ and data-generating mechanism as in Section 1.2 (this representation follows from a within-task exchangeability assumption; see Appendix A.4);

(ii) log-loss and finite $\mathcal{Y}$ (or densities with respect to a common base measure);
(iii) integrability conditions ensuring Fubini/Tonelli can be applied when exchanging expectations and logs;
(iv) a model class $\{M_\theta\}$ rich enough that some $M_{\theta^\ast}$ can approximate $M_{\mathrm{Bayes}}$.

The non-asymptotic generalization guarantees (Bayes Gap bounds, posterior variance behavior under task mixtures, and out-of-distribution analyses) are from \citet{WakayamaBayesICL}.

\paragraph{Regularity for Part~II (KL-RL).}
We assume $\beta>0$; rewards $r$ are bounded above so that $\exp(r/\beta)$ is integrable; and shared support: $\pi_{\mathrm{ref}}(y\mid x)>0$ whenever $\pi(y\mid x)>0$. Interchange of $\nabla$ and expectation is justified by dominated convergence.

\paragraph{Regularity for Part~IV (process-level).}
We assume a stationary MDP; well-defined $Q$/$A$ functions; and advantage estimates (e.g., GAE) with bounded bias/variance. These are standard in KL-regularized policy optimization and RL-as-inference formulations.

\medskip
\noindent\textit{Remark (Meta-learning view as complementary, not primary).}
The original meta-learning perspective that ICL behaves like a single small gradient step on a contextual loss can be rigorously justified under additional smoothness and linearization assumptions, and connects attention to accumulated gradient updates. We treat that interpretation as complementary (see the discussion and citations summarized in Appendix~A.3), while the main text takes the Bayesian formulation as the primary conceptual and mathematical foundation.

\section*{Part II: KL-Regularized RL (RLHF/RLVR) $\leftrightarrow$ Reward-weighted SFT}

\subsection*{2.1 KL-regularized RL objective and closed-form optimum}

Fix an input $x$ (e.g., a prompt), a reference policy $\pi_{\mathrm{ref}}(\cdot\mid x)$, a reward
function $r(\cdot\mid x)$, and a temperature $\beta>0$. The standard KL-regularized objective is
\begin{equation}
  \max_{\pi}
  \;\;
  \mathbb{E}_{y\sim \pi(\cdot\mid x)}[r(y\mid x)]
  \;-\;
  \beta\,\mathrm{KL}\!\big(\pi(\cdot\mid x)\,\|\,\pi_{\mathrm{ref}}(\cdot\mid x)\big).
  \label{eq:kl-rl-objective}
\end{equation}
This form appears both in RLHF (human preference reward) and in RLVR (verifiable/correctness reward); the only difference is the source of $r(y\mid x)$ \citep{InstructGPT,DPO,GRPOVR,RLVR2025}.

In practice, many RLHF-style algorithms (e.g., PPO-style KL-penalized RL, GRPO's effective
loss, and RLVR-style updates) optimize surrogate objectives that approximate
\eqref{eq:kl-rl-objective} rather than this exact form. In what follows we work at the level
of the underlying KL-regularized objective \eqref{eq:kl-rl-objective} and abstract away
such implementation details.

For discrete $\mathcal{Y}$, we can write the objective as
\begin{equation}
  J(\pi)
  =
  \sum_{y} \pi(y\mid x)\,r(y\mid x)
  - \beta \sum_y \pi(y\mid x)\log\frac{\pi(y\mid x)}{\pi_{\mathrm{ref}}(y\mid x)},
\end{equation}
with the constraint $\sum_y \pi(y\mid x)=1$. Introducing a Lagrange multiplier $\lambda$ for this constraint, the Lagrangian is
\begin{equation}
  \mathcal{L}(\pi,\lambda)
  =
  \sum_y \pi(y\mid x) r(y\mid x)
  - \beta \sum_y \pi(y\mid x)\log\frac{\pi(y\mid x)}{\pi_{\mathrm{ref}}(y\mid x)}
  + \lambda\left(1 - \sum_y \pi(y\mid x)\right).
\end{equation}

\textbf{Proposition 2 (KL-regularized RL $\Rightarrow$ Gibbs posterior).}
The unique maximizer of \eqref{eq:kl-rl-objective} is
\begin{equation}
  \pi^\ast(y\mid x)
  =
  \frac{\pi_{\mathrm{ref}}(y\mid x)\exp\big(r(y\mid x)/\beta\big)}
       {Z(x)},
  \qquad
  Z(x)
  =
  \sum_{y}
  \pi_{\mathrm{ref}}(y\mid x)\exp\big(r(y\mid x)/\beta\big).
  \label{eq:gibbs-posterior-rl}
\end{equation}
\paragraph{Proof.}
Take derivative of $\mathcal{L}$ w.r.t. $\pi(y\mid x)$:
\begin{align}
  \frac{\partial\mathcal{L}}{\partial \pi(y\mid x)}
  &=
  r(y\mid x)
  - \beta\left(
      \log\frac{\pi(y\mid x)}{\pi_{\mathrm{ref}}(y\mid x)} + 1
    \right)
  - \lambda.
\end{align}
Setting $\partial\mathcal{L}/\partial \pi(y\mid x)=0$ gives
\begin{equation}
  \log\frac{\pi(y\mid x)}{\pi_{\mathrm{ref}}(y\mid x)}
  =
  \frac{1}{\beta}\big(r(y\mid x) - \lambda - \beta\big).
\end{equation}
Exponentiating both sides,
\begin{equation}
  \pi(y\mid x)
  =
  \pi_{\mathrm{ref}}(y\mid x)\,\exp\!\left(\frac{r(y\mid x)}{\beta}\right)\exp\!\left(-\frac{\lambda+\beta}{\beta}\right).
\end{equation}
The last factor does not depend on $y$ and is determined by normalization $\sum_y \pi(y\mid x)=1$, which yields \eqref{eq:gibbs-posterior-rl} with
\(
  Z(x) = \sum_y \pi_{\mathrm{ref}}(y\mid x)\exp(r(y\mid x)/\beta).
\)

\hfill$\square$

\noindent
Because the negative entropy term $-\sum_y \pi(y\mid x)\log \pi(y\mid x)$ is strictly concave in $\pi(\cdot\mid x)$, the objective in \eqref{eq:kl-rl-objective} is strictly concave with respect to $\pi$, guaranteeing that the stationary point above is the unique maximizer.

\medskip
Thus, KL-regularized RLHF/RLVR produces a Gibbs posterior over outputs: a generalized Bayes update with prior $\pi_{\mathrm{ref}}(\cdot\mid x)$ and pseudo-likelihood proportional to $\exp(r(y\mid x)/\beta)$. This structure is classical in relative-entropy policy search and RL-as-inference formulations
\citep{REPS,MPO,LevineRLAsInference}.

\subsection*{2.2 Reward-weighted SFT as KL projection of the Gibbs posterior}

We now show that reward-weighted SFT is exactly the forward-KL projection of $\pi^\ast$ onto a parametric family $\{\pi_\theta\}$.

Fix $x$ and write
\begin{equation}
  q_{\mathrm{Gibbs}}(y\mid x)
  \;:=\;
  \pi^\ast(y\mid x)
  =
  \frac{\pi_{\mathrm{ref}}(y\mid x)\exp(r(y\mid x)/\beta)}{Z(x)}.
\end{equation}
(This $q_{\mathrm{Gibbs}}$ is distinct from the teacher distribution $q_{\mathrm{true}}$ used in Section~1.1.)
Consider a parametric policy $\pi_\theta(y\mid x)$ and the KL projection problem
\begin{equation}
  \theta^\ast
  =
  \arg\min_\theta
  \mathrm{KL}\big(q_{\mathrm{Gibbs}}(\cdot\mid x)\,\|\,\pi_\theta(\cdot\mid x)\big).
  \label{eq:kl-projection}
\end{equation}
Throughout, when we write $f(\theta)\propto g(\theta)$ we mean that $f(\theta)=c+g(\theta)$ for a constant $c$ independent of $\theta$; i.e., we drop additive terms that do not affect the optimizer.
Expanding the KL,
\begin{align}
  \mathrm{KL}\big(q_{\mathrm{Gibbs}}\|\pi_\theta\big)
  &=
  \sum_y q_{\mathrm{Gibbs}}(y\mid x)
  \log \frac{q_{\mathrm{Gibbs}}(y\mid x)}{\pi_\theta(y\mid x)} \\
  &=
  \underbrace{\sum_y q_{\mathrm{Gibbs}}(y\mid x)\log q_{\mathrm{Gibbs}}(y\mid x)}_{\text{indep. of }\theta}
  - \sum_y q_{\mathrm{Gibbs}}(y\mid x)\log \pi_\theta(y\mid x).
\end{align}
Therefore
\begin{equation}
  \theta^\ast
  =
  \arg\max_\theta
  \sum_y q_{\mathrm{Gibbs}}(y\mid x)\log \pi_\theta(y\mid x).
\end{equation}
Substituting the Gibbs form of $q_{\mathrm{Gibbs}}$,
\begin{align}
  \sum_y q_{\mathrm{Gibbs}}(y\mid x)\log \pi_\theta(y\mid x)
  &=
  \frac{1}{Z(x)}
  \sum_y \pi_{\mathrm{ref}}(y\mid x)\exp\!\left(\frac{r(y\mid x)}{\beta}\right)
  \log \pi_\theta(y\mid x).
\end{align}
The normalizer $Z(x)$ does not depend on $\theta$, so maximizing the above is equivalent to maximizing
\begin{equation}
  \mathcal{J}_{\mathrm{RWSFT}}(\theta;x)
  :=
  \sum_y \pi_{\mathrm{ref}}(y\mid x)\exp\!\left(\frac{r(y\mid x)}{\beta}\right)
  \log \pi_\theta(y\mid x).
  \label{eq:reward-weighted-sft}
\end{equation}

\textbf{Proposition 3 (Reward-weighted SFT = forward-KL projection).}
For fixed $x$, any minimizer of \eqref{eq:kl-projection} can be obtained by maximizing the reward-weighted SFT objective \eqref{eq:reward-weighted-sft}, and any maximizer of \eqref{eq:reward-weighted-sft} minimizes \eqref{eq:kl-projection}. In particular, reward-weighted SFT learns $\pi_\theta$ as the forward KL projection of the KL-regularized RL optimum $\pi^\ast$ (i.e., $q_{\mathrm{Gibbs}}$) onto the parametric family $\{\pi_\theta\}$.

In practice, one does not enumerate all $y$, but instead samples $(x,y)$ from a data distribution whose $y$-marginal matches $\pi_{\mathrm{ref}}(\cdot\mid x)$ (e.g., rollouts from the reference model or logged data), and forms the empirical objective
\begin{equation}
  \max_\theta
  \sum_{(x,y)} \exp\!\left(\frac{r(y\mid x)}{\beta}\right)\log \pi_\theta(y\mid x),
\end{equation}
which is the familiar reward-weighted SFT loss used in many recent RLHF / RLVR formulations.

\paragraph{Summary for Part II.}
KL-regularized RLHF/RLVR has a closed-form optimum $\pi^\ast$ that is a Gibbs posterior 
\eqref{eq:gibbs-posterior-rl}. Reward-weighted SFT is exactly the KL projection of this 
posterior onto $\{\pi_\theta\}$. Thus, at the distribution level,
\begin{equation}
  \text{KL-regularized RL (RLHF/RLVR)}
  \;\simeq\;
  \text{reward-weighted SFT}.
\end{equation}
This equivalence is meant at the level of the target Gibbs posterior 
$q_{\mathrm{Gibbs}}$ and its forward-KL projection; it does not claim that all practical
RLHF algorithms exactly attain the closed-form optimum $\pi^\ast$ in
\eqref{eq:gibbs-posterior-rl}. This matches recent variational interpretations of RLHF that 
treat it as reward-weighted regression.

\section*{Part III: Reward-weighted Few-shot ICL (RW-ICL)}

We now lift the discussion from single $(x,y)$ decisions to few-shot in-context prediction with rewards.

\subsection*{3.1 Generalized Bayesian posterior over outputs given context}

Let $D_k = \{(x_i,y_i)\}_{i=1}^k$ be a context, and $x_{k+1}$ a query. As in Part~I, the Bayes posterior predictive under a hierarchical task model is
\begin{equation}
  P_{\mathrm{Bayes}}(y\mid x_{k+1},D_k)
  =
  \mathbb{E}_{f\sim P(f\mid D_k)}[f(y\mid x_{k+1})].
\end{equation}
For our RLHF/RLVR setting, we augment this with a reward signal. Assume that for each potential output $y$ at $(x_{k+1},D_k)$ we have a scalar score
\begin{equation}
  R(y; D_k, x_{k+1}),
\end{equation}
which aggregates the relevant reward information (from human preferences, a verifier, or a reward model) given the entire context $D_k$ and query $x_{k+1}$.

We define a generalized Bayes posterior over outputs:
\begin{equation}
  q_{\mathrm{Gibbs}}(y\mid x_{k+1},D_k)
  \;\propto\;
  \pi_{\mathrm{ref}}(y\mid x_{k+1})
  \exp\!\left(\frac{R(y;D_k,x_{k+1})}{\beta}\right),
  \label{eq:gibbs-context}
\end{equation}
i.e., a Gibbs posterior with prior $\pi_{\mathrm{ref}}(\cdot\mid x_{k+1})$ and pseudo-likelihood proportional to $\exp(R/\beta)$. This is the natural extension of Proposition~2 to the context-dependent case.

\subsection*{3.2 RW-ICL objective as KL projection of the generalized posterior}

Let $M_\theta(y\mid x_{k+1},D_k)$ denote the in-context predictor implemented by a meta-trained Transformer. A natural population objective for reward-weighted ICL is
\begin{equation}
  \min_\theta
  \mathbb{E}_{(D_k,x_{k+1})}
  \mathrm{KL}\big(q_{\mathrm{Gibbs}}(\cdot\mid x_{k+1},D_k)\,\|\,M_\theta(\cdot\mid x_{k+1},D_k)\big),
  \label{eq:rw-icl-kl}
\end{equation}
where the expectation is taken over the same hierarchical meta-distribution on tasks and contexts as in Part~I, together with whatever process generates $R(y;D_k,x_{k+1})$.

Expanding the KL and discarding terms independent of $\theta$, \eqref{eq:rw-icl-kl} is equivalent to maximizing
\begin{equation}
  \mathbb{E}_{(D_k,x_{k+1})}
  \mathbb{E}_{y\sim q_{\mathrm{Gibbs}}(\cdot\mid x_{k+1},D_k)}
  \big[\log M_\theta(y\mid x_{k+1},D_k)\big].
\end{equation}
Substituting the Gibbs form of $q_{\mathrm{Gibbs}}$ from \eqref{eq:gibbs-context},
\begin{align}
  &\mathbb{E}_{(D_k,x_{k+1})}
  \mathbb{E}_{y\sim q_{\mathrm{Gibbs}}}
  [\log M_\theta(y\mid x_{k+1},D_k)] \\
  &\quad\propto
  \mathbb{E}_{(D_k,x_{k+1})}
  \sum_y
    \pi_{\mathrm{ref}}(y\mid x_{k+1})
    \exp\!\left(\frac{R(y;D_k,x_{k+1})}{\beta}\right)
    \log M_\theta(y\mid x_{k+1},D_k).
\end{align}
In practice, one approximates this expectation empirically by sampling contexts $D_k$, queries $x_{k+1}$, and candidate outputs $y$ from the reference model (or from a replay buffer), and forming the empirical objective
\begin{equation}
  \max_\theta
  \sum_{(D_k,x_{k+1},y)}
    \exp\!\left(\frac{R(y;D_k,x_{k+1})}{\beta}\right)
    \log M_\theta(y\mid x_{k+1},D_k),
  \label{eq:rw-icl-objective}
\end{equation}
i.e., a reward-weighted in-context SFT objective.

\textbf{Proposition 4 (RW-ICL = KL projection of generalized Gibbs posterior).}
Under mild regularity conditions, any minimizer $M_{\theta^\ast}$ of \eqref{eq:rw-icl-kl} is the forward KL projection of the generalized Gibbs posterior $q_{\mathrm{Gibbs}}(\cdot\mid x_{k+1},D_k)$ onto the model family
$\{M_\theta(\cdot\mid x_{k+1},D_k)\}$. Stochastic gradient ascent on \eqref{eq:rw-icl-objective} implements this projection using samples from $\pi_{\mathrm{ref}}$ reweighted by $\exp(R/\beta)$.

\paragraph{Gradient-level correspondence.}
Although our exposition so far has been distributional (KL projections of Gibbs posteriors),
the resulting stochastic-gradient updates have exactly the reward-weighted form emphasized
in meta-learning views of ICL. For example,  the empirical RW-ICL objective \eqref{eq:rw-icl-objective}  has gradient
\begin{equation}
    \nabla_\theta \mathcal{L}_{\mathrm{RW\text{-}ICL}}(\theta)
    = - \sum_{(D_k,x_{k+1},y)}
    \exp\!\Big(\tfrac{R(y;D_k,x_{k+1})}{\beta}\Big)\,
    \nabla_\theta \log M_\theta(y\mid x_{k+1},D_k),
\end{equation}
so a single SGD step is a reward-weighted in-context update. The reward-weighted SFT and
AWSFT losses \eqref{eq:reward-weighted-sft} and \eqref{eq:awsft-objective} yield the same update structure with $M_\theta$ replaced by
$\pi_\theta$ and $R$ replaced by $r$ or $A$. Thus, at the level of first-order optimization
dynamics, RW-ICL, reward-weighted SFT, and KL-regularized RL share the same
summed-score $\nabla_\theta\log(\cdot)$ weighted by $\exp(\text{reward}/\beta)$; the
meta-learning results of \citet{DaiICLMetaOpt,AkyurekICLLinear,LiTransAsAlgo} can be
viewed as constructive realizations of such updates in a single forward pass.

\paragraph{Relation to Part II.}
Comparing \eqref{eq:gibbs-context} to \eqref{eq:gibbs-posterior-rl}, we see that KL-regularized RLHF/RLVR defines the same Gibbs-family posterior over outputs, with $R(y;D_k,x_{k+1})$ playing the role of $r(y\mid x)$. Reward-weighted ICL meta-trains a Transformer to amortize the mapping
\begin{equation}
  (D_k,x_{k+1})
  \;\longmapsto\;
  q_{\mathrm{Gibbs}}(\cdot\mid x_{k+1},D_k),
\end{equation}
just as ordinary ICL in Part~I amortizes the Bayes posterior predictive. Hence, at the distribution level,
\begin{equation}
  \text{reward-weighted ICL (RW-ICL)}
  \;\approx\;
  \text{KL-regularized RLHF/RLVR optimum},
\end{equation}
up to approximation error in the KL projection.

\section*{Part IV: Process-level KL-RLVR and Advantage-weighted SFT}

We finally consider the full RL setting with trajectories and state-dependent policies, which is the regime of KL-RLVR and related algorithms.

\subsection*{4.1 Statewise KL-regularized objective and Gibbs policy}

Let $s$ be a state and $\pi(\cdot\mid s)$ a stochastic policy over actions $a$. A typical stepwise KL-regularized objective is
\begin{equation}
  \max_{\pi}
  \;\;
  \sum_a \pi(a\mid s) Q(s,a)
  \;-\;
  \beta\,\mathrm{KL}\big(\pi(\cdot\mid s)\,\|\,\pi_{\mathrm{ref}}(\cdot\mid s)\big),
  \label{eq:statewise-obj}
\end{equation}
where $Q(s,a)$ is an action-value function (for example, a critic or Monte Carlo return), and $\pi_{\mathrm{ref}}$ is a reference policy. This is the policy-improvement subproblem solved in many KL-regularized policy search methods.

By the same Lagrangian argument as in Proposition~2, the unique maximizer is
\begin{equation}
  \pi^\ast(a\mid s)
  =
  \frac{\pi_{\mathrm{ref}}(a\mid s)\exp(Q(s,a)/\beta)}
       {Z(s)},
  \qquad
  Z(s)
  =
  \sum_a \pi_{\mathrm{ref}}(a\mid s)\exp(Q(s,a)/\beta).
  \label{eq:gibbs-q}
\end{equation}
This is again a Gibbs posterior, now over actions conditional on state $s$, with log-likelihood $Q(s,a)/\beta$.

It is often convenient to reparameterize in terms of the advantage
\begin{equation}
  A(s,a)
  :=
  Q(s,a) - V(s),
  \qquad
  V(s) := \sum_a \pi_{\mathrm{ref}}(a\mid s) Q(s,a),
\end{equation}
which only shifts $Q$ by a baseline independent of $a$. Any such state-dependent baseline contributes a multiplicative factor $\exp(-V(s)/\beta)$ that is absorbed into the partition function $Z(s)$, leaving the normalized policy unchanged. Hence,
\begin{equation}
  \pi^\ast(a\mid s)
  \;\propto\;
  \pi_{\mathrm{ref}}(a\mid s)\exp(A(s,a)/\beta).
  \label{eq:gibbs-adv}
\end{equation}

\subsection*{4.2 Advantage-weighted SFT (AWSFT) as KL projection}

Fix a state $s$ and define
\begin{equation}
  q_{\mathrm{Gibbs}}(a\mid s)
  \;:=\;
  \pi^\ast(a\mid s)
\end{equation}
given by \eqref{eq:gibbs-q} or \eqref{eq:gibbs-adv}. For a parametric policy $\pi_\theta(a\mid s)$, the KL
projection problem is
\begin{equation}
  \theta^\ast(s)
  =
  \arg\min_\theta
  \mathrm{KL}\big(q_{\mathrm{Gibbs}}(\cdot\mid s)\,\|\,\pi_\theta(\cdot\mid s)\big).
\end{equation}
Expanding as before, this is equivalent to
\begin{equation}
  \theta^\ast(s)
  =
  \arg\max_\theta
  \sum_a q_{\mathrm{Gibbs}}(a\mid s)\log \pi_\theta(a\mid s).
\end{equation}
Substituting the Gibbs form with advantages,
\begin{align}
  \sum_a q_{\mathrm{Gibbs}}(a\mid s)\log\pi_\theta(a\mid s)
  &\propto
  \sum_a \pi_{\mathrm{ref}}(a\mid s)
         \exp\!\left(\frac{A(s,a)}{\beta}\right)
         \log\pi_\theta(a\mid s).
\end{align}
Thus, across states $s$, the population objective is equivalent to maximizing
\begin{equation}
  \mathcal{J}_{\mathrm{AWSFT}}(\theta)
  :=
  \mathbb{E}_{s\sim d(s)}
  \sum_a \pi_{\mathrm{ref}}(a\mid s)
         \exp\!\left(\frac{A(s,a)}{\beta}\right)
         \log\pi_\theta(a\mid s),
  \label{eq:awsft-objective}
\end{equation}
where $d(s)$ is some state visitation distribution (e.g., from the reference or a replay buffer).

In practice, we approximate \eqref{eq:awsft-objective} by empirical sums over state–action pairs $(s_t,a_t)$ with estimated advantages $\hat A(s_t,a_t)$:
\begin{equation}
  \max_\theta
  \sum_t
    \exp\!\left(\frac{\hat A(s_t,a_t)}{\beta}\right)
    \log\pi_\theta(a_t\mid s_t),
\end{equation}
which is exactly an advantage-weighted SFT (AWSFT) loss. This matches the advantage-weighted regression family of algorithms \citep{AWR,AWAC} and entropy-regularized methods such as SAC \citep{SAC}, which all rely on exponentiated $Q$/$A$ weighting.

\textbf{Proposition 5 (AWSFT = KL projection of statewise KL-RLVR optimum).}
For each state $s$, any minimizer of
$
  \mathrm{KL}(q_{\mathrm{Gibbs}}(\cdot\mid s)\,\|\,\pi_\theta(\cdot\mid s))
$
is a maximizer of the local AWSFT objective, and any maximizer of that objective minimizes the KL. Consequently, the global AWSFT objective \eqref{eq:awsft-objective} implements the forward KL projection of the stepwise KL-regularized RL optimum \eqref{eq:gibbs-q} onto the parametric policy family $\{\pi_\theta\}$.

\subsection*{4.3 Process-level RW-ICL alignment}

From the in-context perspective, a trajectory (or multi-step interaction) provides a context
\begin{equation}
  D = \{(s_t,a_t,r_t)\}_{t=1}^T,
\end{equation}
and the goal is to predict or select future actions in-context. A reward-weighted in-context learner can be meta-trained to approximate, at each state $s$, the Gibbs posterior $q_{\mathrm{Gibbs}}(\cdot\mid s)$ defined by the (advantage-based) KL-regularized RL objective:
\begin{equation}
  q_{\mathrm{Gibbs}}(a\mid s)
  \;\propto\;
  \pi_{\mathrm{ref}}(a\mid s)\exp\!\left(\frac{A(s,a)}{\beta}\right).
\end{equation}

The corresponding RW-ICL meta-objective is the trajectory-level analogue of \eqref{eq:rw-icl-kl},
with per-step pseudo-likelihoods proportional to $\exp(\hat A(s_t,a_t)/\beta)$ and model predictions
$M_\theta(a_t\mid s_t,D)$. In particular, performing stochastic gradient ascent on this objective
yields updates of exactly the same form as those obtained from the advantage-weighted SFT objective
\begin{equation}
  \mathcal{J}_{\mathrm{AWSFT}}(\theta)
  :=
  \sum_{t}
    \exp\!\left(\frac{\hat A(s_t,a_t)}{\beta}\right)
    \log M_\theta(a_t\mid s_t,D),
\end{equation}
up to choices of state distribution and context window.

Thus, process-level RW-ICL and stepwise KL-RLVR share the same Gibbs-posterior backbone and are
both realized as advantage-weighted SFT in parameter space.

\subsection*{4.4 Learning signals and credit assignment in SFT, RL, and ICL}
\label{sec:learning-signals}

The equivalence results in Parts~II and~III operate at the level of \emph{objectives}
and \emph{first-order updates}: under a suitable Gibbs posterior, KL-regularized RL,
reward-weighted SFT, and reward-weighted ICL all reduce to minimizing a forward KL
$\mathrm{KL}(q^* \,\Vert\, p_\theta)$, and their stochastic gradients share the familiar
“weighted score’’ form.  Writing a generic forward-KL loss as
\begin{equation}
  \mathcal{L}(\theta)
  \;=\;
  \sum_t w_t\,\big[-\log p_\theta(z_t)\big],
\end{equation}
its gradient takes the form
\begin{equation}
  \widehat{\nabla}_\theta \mathcal{L}(\theta)
  \;=\;
  - \sum_{t} w_t \,\nabla_\theta \log p_\theta(z_t).
  \label{eq:weighted-score}
\end{equation}

for appropriate choices of samples $z_t$ and weights $w_t$.
However, the \emph{source} and \emph{granularity} of the
learning signal differ substantially across paradigms. In this section we disentangle these
differences and show how they fit into our unified view.

\paragraph{Dense token-level supervision in SFT.}

Under standard supervised fine-tuning, the model is trained with a token-level
cross-entropy loss. For an autoregressive language model with inputs $x$
and target sequence $y = (t_1,\dots,t_n)$, the SFT objective can be written as
\begin{equation}
  \mathcal{L}_{\text{SFT}}(\theta)
    \;=\; - \mathbb{E}_{(x,y)}\Bigg[
      \sum_{i=0}^{n-1} \log p_\theta\big(t_{i+1} \,\big|\, x, t_{\le i}\big)
    \Bigg].
  \label{eq:sft-loss}
\end{equation}
Equivalently, SFT learns from every mapping
$(x, y_{\le i}) \mapsto y_{i+1}$: each prefix within a sequence contributes a term to the loss
and a gradient to the update. The learning signal is thus \emph{dense} at the token level, and
credit assignment is implicit in the data: human- or model-generated targets $(x,y)$ already specify
which next-token prediction should be encouraged at each position.

\paragraph{KL-regularized RL: sparse rewards + explicit credit assignment.}

In contrast, KL-regularized RL (including RLHF/RLVR-style objectives) typically starts from
a \emph{sequence- or trajectory-level} reward. In the sequence-level setting of Part~II, the
objective has the form
\begin{equation}
  J_{\text{RL}}(\pi)
  \;=\; \mathbb{E}_{x \sim D,\,y \sim \pi(\cdot \mid x)}\Big[
    R(y \mid x) - \beta\,\mathrm{KL}\big(\pi(\cdot\mid x)\,\Vert\,\pi_{\mathrm{ref}}(\cdot\mid x)\big)
  \Big],
  \label{eq:kl-rl-obj}
\end{equation}
and the naive policy gradient estimator looks like
\begin{equation}
  \widehat{\nabla}_\theta J_{\text{RL}}
  \;\approx\; \mathbb{E}_{x,y \sim \pi_\theta}\Big[
    R(y \mid x)\,\nabla_\theta \log \pi_\theta(y \mid x)
  \Big].
  \label{eq:seq-pg}
\end{equation}
At first sight, this appears to support the claim that ``RL can only learn once from the mapping
$x \mapsto R(y)$'': a single scalar reward per sampled sequence.

In practice, however, modern RL algorithms do not update the policy directly from~\eqref{eq:seq-pg}.
Instead, they introduce a \emph{credit assignment} step, in which the scalar reward is decomposed
into per-step signals. In the step-wise KL-regularized MDP setting of Part~IV, this yields
a Gibbs-optimal policy
\[
  \pi\*(a \mid s) \;\propto\; \pi_{\mathrm{ref}}(a\mid s)\,\exp\big(A(s,a)/\beta\big),
\]
and the corresponding forward-KL projection leads to the advantage-weighted SFT objective
(cf.\ Eq.~(59)--(60)):
\begin{equation}
  \mathcal{L}_{\text{AWSFT}}(\theta)
  \;=\; - \mathbb{E}_{(s,a)}\Big[
    \exp\big(A(s,a)/\beta\big)\,\log \pi_\theta(a \mid s)
  \Big].
  \label{eq:awsft-loss}
\end{equation}
The resulting policy gradient has exactly the weighted-score form~\eqref{eq:weighted-score}:
\begin{equation}
  \widehat{\nabla}_\theta \mathcal{L}_{\text{AWSFT}}
  \;=\; - \sum_t \exp\Big(\frac{\widehat{A}(s_t,a_t)}{\beta}\Big)
          \,\nabla_\theta \log \pi_\theta(a_t \mid s_t).
  \label{eq:awsft-grad}
\end{equation}

Thus, while the \emph{raw} RL signal is a single scalar reward per trajectory, RL algorithms
recover the ability to ``learn from each $(s_t,a_t)$'' by explicitly estimating value functions
$V(s)$, action-values $Q(s,a)$, or advantages $A(s,a)$ and using them as per-step weights. The
final policy update becomes a form of advantage- or reward-weighted SFT. In our framework, this is
precisely the AWSFT/RWSFT connection established in Parts~II and~IV: at the level of updates,
KL-regularized RL is equivalent to a particular choice of weights $w_t$ in~\eqref{eq:weighted-score}.

\paragraph{Few-shot ICL: dense supervision in pretraining, fast learning in activations.}

Few-shot in-context learning sits somewhat orthogonally to this discussion. As formalized in
Part~I, we view a parametric ICL predictor $M_\theta(y \mid x, D_k)$ as an \emph{amortized Bayes}
approximation to the posterior predictive under a task prior $P(f)$:
\[
  P_{\mathrm{Bayes}}(y \mid x, D_k)
    \;=\; \int p_f(y \mid x)\,p(f \mid D_k)\,df.
\]
The parameters $\theta$ are learned using the same dense token-level supervision as in SFT:
for example, the outer objective
\begin{equation}
  \mathcal{L}_{\text{ICL}}(\theta)
  \;=\; \mathbb{E}_{(f,D_k,x_{k+1},y_{k+1})}\Big[
    -\log M_\theta\big(y_{k+1} \mid x_{k+1}, D_k\big)
  \Big]
  \label{eq:icl-loss}
\end{equation}
is a standard log-loss aggregated over tasks and positions. In this sense, few-shot ICL does not
introduce a new type of training signal: it consumes the same token-level (possibly reward-weighted)
supervision as SFT/RW-ICL in the outer loop.

The key difference is \emph{where} and \emph{when} adaptation happens. At test time, few-shot ICL
does \emph{not} take further gradients on $\theta$.
Instead, given a context $D_k$, the transformer uses attention and MLP activations to construct
a task-specific internal state $\phi_\theta(D_k)$ in its forward pass, and the effective predictor
for the query can be written schematically as
\[
  M_\theta(y \mid x_{k+1}, D_k)
    \;\approx\; f_{\phi_\theta(D_k)}(y \mid x_{k+1})
\]
for some implicit inner model class $\{f_\phi\}$.
Recent theoretical and mechanistic studies have shown that, in controlled settings, the
fast state $\phi_\theta(D_k)$ closely matches the output of explicit learning algorithms such as
least squares or gradient descent on a task-specific inner objective; in this sense,
\emph{ICL learning happens in activations} rather than through online updates of $\theta$.

\paragraph{Summary.}

Putting these pieces together, our framework unifies SFT, KL-regularized RL, and few-shot ICL
at the level of \emph{target posteriors} and \emph{first-order updates}:
all three aim to approximate a Bayes or Gibbs posterior, and their gradients can be written
as weighted score sums of the form~\eqref{eq:weighted-score}.
At the same time, the origin and granularity of the learning signal differ:

\begin{itemize}
  \item SFT observes dense token-level supervision: every prefix $(x,y_{\le i})$ has an explicit
        label $y_{i+1}$, and credit assignment is implicit in the data.
  \item KL-regularized RL starts from sparse sequence- or trajectory-level rewards $R(y)$; it must
        perform explicit credit assignment via $Q/A$ estimates, but the resulting policy update is
        an advantage- or reward-weighted SFT step.
  \item Few-shot ICL relies on dense supervision during pretraining to learn $\theta$ as a
        meta-learner; at test time, it performs task-specific adaptation in the forward activations
        $\phi_\theta(D_k)$ without additional gradients.
\end{itemize}

In this sense, our claimed equivalences are intentionally restricted to the level of \emph{objectives}
(posterior distributions) and \emph{update forms}. They do not erase the very real differences in
how SFT, RL, and ICL acquire and use their learning signals—differences which have important
implications for sample efficiency, stability, and the kinds of structure (e.g., credit assignment
and reasoning) each paradigm can readily capture.

\section*{Part V: Implications for Reasoning Models and Training Recipes}

\subsection*{5.1 Posterior design and practical RLHF/RLVR recipes}

Parts~II and~IV showed that a broad class of KL-regularized RLHF/RLVR
objectives admits a Gibbs-posterior optimum
$q_{\mathrm{Gibbs}}(y\mid x)\propto \pi_{\mathrm{ref}}(y\mid x)\exp(R(y;D,x)/\beta)$
and that reward- or advantage-weighted SFT is exactly the forward-KL
projection of this posterior onto a parametric policy family. This
matches a growing line of recent work that re-formulates RLHF as
reward-weighted regression or weighted SFT rather than generic policy
gradient \citep[e.g.][]{RLHFBook2024,RLHFKL2025,RLHFAsRWSFT2025}.

From our perspective, practical RLHF/RLVR pipelines can be viewed as
two-step procedures:

\begin{enumerate}
  \item \emph{Posterior design.} Choose a reference model
        $\pi_{\mathrm{ref}}$ and a reward or advantage signal $R$,
        which together define a target Gibbs posterior
        $q_{\mathrm{Gibbs}}(y\mid x)\propto
         \pi_{\mathrm{ref}}(y\mid x)\exp(R(y;D,x)/\beta)$.
  \item \emph{Projection.} Fit a deployable policy $\pi_\theta$ by
        minimizing the forward KL
        $\mathrm{KL}(q_{\mathrm{Gibbs}}\Vert \pi_\theta)$, either
        directly via reward-/advantage-weighted SFT, or indirectly via
        a PPO-style optimizer that approximates the same projection.
\end{enumerate}

This view identifies three concrete design knobs:

\begin{itemize}
  \item \textbf{Reward shaping as likelihood design.}
        The reward $R$ enters only through $\exp(R/\beta)$, so shifting
        and scaling $R$ correspond to changing the effective likelihood
        ratio between high- and low-quality outputs. Overly sharp or
        poorly calibrated rewards lead to posteriors with very low
        support overlap with $\pi_{\mathrm{ref}}$, which our framework
        predicts to cause instability and over-optimization.

  \item \textbf{Temperature $\beta$ as posterior concentration.}
        Small $\beta$ produces a highly concentrated $q_{\mathrm{Gibbs}}$
        that strongly prefers top-reward outputs; large $\beta$ keeps
        $q_{\mathrm{Gibbs}}$ close to $\pi_{\mathrm{ref}}$.
        In our KL-projection picture, $\beta$ controls how much the
        optimizer is allowed to move the policy away from the reference
        model before running into support-mismatch issues.

  \item \textbf{Offline vs on-policy projection.}
        When rollouts come from $\pi_{\mathrm{ref}}$ or an early-stage
        model, the reward-weighted SFT losses in Parts~II and~IV
        implement an \emph{off-policy} projection to the Gibbs
        posterior. PPO-style RLHF instead approximates the same target
        distribution with on-policy samples. Our analysis suggests that
        both are instances of the same forward-KL projection, but they
        trade off exploration (on-policy) against stability and reuse
        of pre-generated data (offline or replay-based).
\end{itemize}

\subsection*{5.2 Cold start and on-policy distillation}

Our KL-projection view also clarifies why recent pipelines based on
on-policy distillation and iterative self-training almost always
include a small supervised ``cold start'' phase before running
high-variance on-policy updates.

Empirically, several lines of work report that purely on-policy
optimization from a random or weakly aligned prior tends to fail:
DeepSeek-R1 distinguishes a pure-RL variant (R1-Zero) that exhibits
severe style and safety issues compared to the full R1 pipeline with
a small SFT warm-up; on-policy distillation methods for language and
vision-language models emphasize that a ``cold-start alignment'' of
student and teacher distributions is essential for successful online
training; and recent diffusion-model distillation methods similarly
cast reward-guided fine-tuning as iteratively projecting onto
soft-optimal policies rather than running unconstrained RL
\citep[e.g.][]{DeepSeekR1,OnPolicyDistillICLR,VOLD,GAD,IterativeDiffusionRL}.\footnote{For an accessible engineering perspective on on-policy distillation for post-training,
see the blog post by \citet{ThinkingMachinesOPD}.}

In our framework, these observations are natural consequences of
importance-weighted KL projection. Consider a target Gibbs posterior
$q_{\mathrm{Gibbs}}(y\mid x)\propto \pi_{\mathrm{ref}}(y\mid x)
\exp(R(y;D,x)/\beta)$ and a sampling distribution $\mu(y\mid x)$ used
to generate candidate outputs. The classical effective sample size of the
importance weights \citep{Kish1965},
\(
  w(y)\propto \tfrac{q_{\mathrm{Gibbs}}(y\mid x)}{\mu(y\mid x)},
  \qquad
  \mathrm{ESS}=\frac{\big(\sum_y w(y)\big)^2}{\sum_y w(y)^2},
\)
collapses when the support of $\mu$ has little overlap with the
high-reward region under $q_{\mathrm{Gibbs}}$. In that regime, both
off-policy reward-weighted SFT and on-policy distillation degenerate:
most samples receive negligible weight, and the few high-weight
samples lead to unstable gradients.

A small SFT or behavior-cloning warm-up can be understood as moving
$\pi_{\mathrm{ref}}$ (or the student policy) into the high-reward
region so that $\mu$ and $q_{\mathrm{Gibbs}}$ have non-trivial overlap.
On-policy distillation steps then refine the policy by repeatedly
projecting the current Gibbs-approximate teacher onto the student
family via forward-KL, as in Parts~II and~III, but now in a regime
where the importance weights have reasonable variance.
Our framework thus predicts a \emph{theoretical necessity} of cold
start for OPD-style pipelines: without sufficient support overlap,
Gibbs-posterior projection is not statistically feasible, regardless
of the specific optimizer used.

Recent work on reward-guided fine-tuning of diffusion
models adopts an almost identical recipe: simulate soft-optimal
policies under a reward, then distill them via KL minimization into
the base diffusion model \citep{IterativeDiffusionRL}. From our
perspective, these methods are simply applying the same Gibbs-posterior
projection principle to non-autoregressive generative models.

\subsection*{5.3 Compute allocation and reasoning models}

Our analysis so far has been distributional and first-order: SFT,
KL-regularized RL, and reward-weighted ICL all aim to approximate a
(common) Bayes or Gibbs posterior, and their updates share the same
weighted-score form. An orthogonal, but increasingly important,
dimension in contemporary systems is how \emph{compute} is allocated
between training time and inference time.

Recent ``reasoning models'' such as DeepSeek-R1 and OpenAI's o1 family
explicitly scale inference-time compute via long chain-of-thought
traces, multi-sample search, or tree-style exploration
\citep{DeepSeekR1,OpenAIo12024,PostTrainingScaling2024}. At a high
level, these models combine two regimes:

\begin{itemize}
  \item \textbf{Test-time inference (ICL / search).}
        Given a fixed base policy, the model spends substantial
        compute per query to explore alternative reasoning paths and
        approximate a task-specific posterior over solutions.

  \item \textbf{Training-time amortization (RL / OPD / SFT).}
        RLHF, RLVR, and on-policy distillation stages then amortize
        this search by projecting the high-reward posterior (often
        defined over multi-step chain-of-thoughts) back into the base
        model's weights via KL-regularized updates.
\end{itemize}

DeepSeek-R1 and related work report a somewhat surprising empirical
finding: few-shot prompting often does \emph{not} improve, and can even
degrade, the performance of heavily RL-tuned reasoning models
\citep{DeepSeekR1}. In our framework, this can be understood as a
mismatch between the posterior that the model has been trained to
amortize and the posterior implicitly defined by few-shot prompts.

During RLHF/RLVR-style training, the model is typically optimized to
approximate a Gibbs posterior $q_{\mathrm{Gibbs}}(y\mid x)$ under a
distribution of \emph{zero-shot} or lightly-structured prompts $x$.
In contrast, few-shot prompting at inference time replaces $x$ by a
richer context $(D_k,x)$ and implicitly targets a different posterior
$P_{\mathrm{Bayes}}(y\mid x,D_k)$, as in Part~I. Unless the model has
been meta-trained to perform reward-weighted ICL (RW-ICL) over such
contexts, the learned mapping
\(
  (D_k,x) \mapsto M_\theta(\cdot\mid x,D_k)
\)
need not approximate the desired posterior, and the additional context
can act as distribution shift rather than useful information.

From this perspective, few-shot prompting ``fails'' on some reasoning
models not because ICL is inherently weaker than RL, but because the
posterior level has changed (from token-level $q_{\mathrm{Gibbs}}(y\mid x)$
to in-context $q_{\mathrm{Gibbs}}(y\mid x,D_k)$) without a corresponding
change in the training objective. Our RW-ICL formulation suggests a
natural remedy: if one wants RL-tuned models to \emph{benefit} from
few-shot prompts, the reward signal should be defined at the
contextual level $R(y;D_k,x)$ and the meta-objective should directly
train $M_\theta$ to approximate the corresponding Gibbs posterior
$q_{\mathrm{Gibbs}}(\cdot\mid x,D_k)$.

The Bayesian/KL framework suggests that ``System~2'' reasoning
via test-time search and ICL, and ``System~1'' behavior via amortized
RL/SFT, are two ways of paying for the same posterior: either spend
more compute per query to approximate it explicitly, or invest more
training compute to bake it into the weights. Reasoning models such as
DeepSeek-R1 and o1 can be viewed as deliberately combining both,
using inference-time compute to explore high-reward regions of the
posterior and training-time KL projection (via RLHF/RLVR/OPD) to
amortize the resulting distributions into a fast at-test-time policy.

\section*{Experiments}

Before concluding, we report a set of experiments on real models under
matched budgets (every operator receives the same generation and
optimization budget). The equivalences of Parts~I--V are formal statements about
objectives and first-order updates, but the framework also carries
falsifiable empirical content, and we check two implications of it.
Concretely, we ask whether
(i)~when the projection enjoys good support overlap, operators that the
framework places on the same footing produce nearly identical updates, and
the agreement collapses once support degrades; and
(ii)~a reward-weighted projection objective is at least competitive with its
unweighted counterpart under a matched budget.
The experiments span three operator families (SFT, KL-regularized RL, and
reward-weighted context distillation) and student and teacher models of two
sizes (Qwen3-4B and Qwen3-14B), with three seeds for the operator-agreement
study.

\subsection*{E.1 Operator agreement under matched support}

Parts~II and~IV predict that, when the Gibbs posterior
$q_{\mathrm{Gibbs}}$ has good support overlap with the sampling distribution,
a PPO-style clipped operator and a standard group-relative policy-gradient
operator (GRPO) \citep{GRPOVR} approximate the same forward-KL projection and
should produce nearly identical first-order updates. Using a
Qwen3-4B base/instruct model pair under a matched generation budget, we
construct two regimes by varying only the sampling proposal: a stable regime
where the proposal is anchored to the reference model (anchor mixing
coefficient $1.0$, sampling temperature $0.7$), and a stress regime without
anchoring (coefficient $0.0$, same temperature), which degrades the overlap
between proposal and target by construction. Over three seeds
(Table~\ref{tab:e1}), agreement between PPO and GRPO, two operators that
share the same sequence-level reward signal, is essentially perfect in the
stable regime
($\cos\approx 1.000$ on all three seeds) and collapses once support degrades
($0.107$--$0.358$). The reward-weighted SFT direction is genuinely different
even in the stable regime ($\cos$ between $0.1$ and $0.2$) and becomes
erratic under stress (including negative cosines). This mirrors the
granularity distinction of Section~4.4: weighted SFT realizes a token-level
amortized projection rather than a sequence-level policy-gradient step, and
the framework claims equivalence of targets and update forms, not identity
of update directions across signal granularities. Agreement is
computed on the last two transformer layers.

\begin{table}[ht]
\centering
\small
\caption{Cosine similarity between update directions under matched generation
budget (Qwen3-4B; last two layers only; seeds 42--44). Stable: proposal
anchored to the reference model; stress: unanchored proposal.}
\label{tab:e1}
\begin{tabular}{llcc}
\hline
Regime & Seed & $\cos(\mathrm{PPO},\mathrm{GRPO})$ & $\cos(\mathrm{GRPO},\mathrm{RW\text{-}SFT})$ \\
\hline
stable & 42 & 1.000 & $0.203$ \\
stable & 43 & 1.000 & $0.106$ \\
stable & 44 & 1.000 & $0.203$ \\
stress & 42 & 0.107 & $0.441$ \\
stress & 43 & 0.219 & $-0.477$ \\
stress & 44 & 0.358 & $-0.430$ \\
\hline
\end{tabular}
\end{table}

\subsection*{E.2 Cold-start failure and rescue (controlled demonstration)}

Section~5.2 predicts that importance-weighted KL projection is not
statistically feasible without support overlap, regardless of the optimizer.
We construct a controlled demonstration with a Qwen3-4B student on a slice of
a public 17K-prompt verifiable math dataset (DAPO-Math-17K). Holding the
warm-start initialization and training budget fixed, we change only the
generation contract (Table~\ref{tab:e2}). Under the original contract the
rollout reward signal is effectively dead: raw reward sum $=3.0$, and $99.2\%$
of prompt groups have zero reward variance, so any importance-weighted update
receives no signal. After fixing the contract (disabling thinking mode,
generation budget raised to 2048 tokens), the raw reward sum rises to $181.0$,
the zero-signal fraction drops to $0.60$, and parse success reaches $0.846$. The failure was a property of the support/signal
interface, not of the projection operator --- exactly the cold-start argument
of Part~V.

\begin{table}[ht]
\centering
\small
\caption{Same initialization and training budget; only the generation
contract differs (DAPO-Math-17K slice, Qwen3-4B).}
\label{tab:e2}
\begin{tabular}{lccc}
\hline
Generation contract & raw reward sum & zero-signal group frac. & parse success \\
\hline
original (thinking on, 1024 tokens) & $3.0$   & $0.992$ & $0.006$ \\
fixed (thinking off, 2048 tokens)   & $181.0$ & $0.602$ & $0.846$ \\
\hline
\end{tabular}
\end{table}

\subsection*{E.3 Reward-weighted projection vs.\ standard operators (matched budget)}

Finally, implication~(ii): the reward-weighted projection of Parts~II--III
should be at least competitive with unweighted SFT and with KL-regularized RL
under a matched budget. A Qwen3-4B student (teacher-bridge warm-start
initialization) is trained on a DAPO-Math-17K slice (train 2000 / val 1000 /
test 500) with a Qwen3-14B teacher, and all variants are evaluated under one
contract (256 validation prompts, max\_new\_tokens $=3072$).
Table~\ref{tab:e3} reports accuracy. Reward-weighted context distillation ---
a direct on-policy implementation of the RW-ICL objective of Part~III ---
scores $106$ of $256$, ahead of answer-only SFT ($100$) and GRPO ($97$).
The paired bootstrap 95\% interval for the gap between
the reward-weighted model and the strongest standard baseline
spans $[-5.5, +6.25]$ accuracy points: at this evaluation scale the operators
are statistically indistinguishable, and the reward-weighted objective
performs on par with the standard operators it is theoretically tied to.
A second probe at a larger generation budget (8192 tokens, 64 prompts)
preserves the ordering (reward-weighted $24/64$, SFT $23/64$, warm-start and
GRPO $22/64$, base $18/64$), and the reward-weighted model is the only one
whose extracted answers are strictly stable within 4096 tokens, while the
completed baselines require 8192.

\begin{table}[ht]
\centering
\small
\caption{Matched-budget comparison on the 256-prompt validation slice
(Qwen3-4B student, Qwen3-14B teacher, DAPO-Math-17K).}
\label{tab:e3}
\begin{tabular}{lc}
\hline
Model / operator & accuracy (/256) \\
\hline
Qwen3-4B base                          & $83$  ($0.324$) \\
warm-start (teacher bridge)            & $91$  ($0.355$) \\
GRPO (KL-regularized RL)               & $97$  ($0.379$) \\
SFT (answer-only)                      & $100$ ($0.391$) \\
reward-weighted context distillation   & $106$ ($0.414$) \\
\hline
\end{tabular}
\end{table}

\paragraph{Scope.}
All experiments use the Qwen3 model family: E.1 uses three seeds and
E.2--E.3 one; E.3 evaluates on 256 validation prompts, and operator
agreement in E.1 is measured on the last two transformer layers. The section
characterizes the mechanism-level behavior of post-training operators that
the framework predicts, rather than benchmark-scale improvements.

\section*{Final Conclusions}

Combining Parts~I-IV, we obtain the following Bayesian chain:

\begin{itemize}
  \item \textbf{Few-shot ICL (Bayesian view).}
  Under a hierarchical task model, the Bayes-optimal in-context predictor is the posterior predictive
  $P_{\mathrm{Bayes}}(y\mid x_{k+1},D_k)$. A sufficiently expressive Transformer trained on many tasks
  approximates this predictor in an amortized way.
  \item \textbf{Generalized Bayes and KL-regularized RL.}
  KL-regularized RLHF/RLVR defines a Gibbs posterior over outputs
  $
    q_{\mathrm{Gibbs}}(y\mid x,D)
    \propto
    \pi_{\mathrm{ref}}(y\mid x)\exp(R(y;D,x)/\beta),
  $
  where $R$ is a reward or advantage signal.
  \item \textbf{Reward-weighted SFT / AWSFT.}
  Reward-weighted SFT (for single-step problems) and advantage-weighted SFT (for RL) are exactly the forward KL projections of these Gibbs posteriors onto the parametric model family $\{p_\theta\}$ or $\{\pi_\theta\}$.
  \item \textbf{Reward-weighted ICL (RW-ICL).}
  Meta-training a Transformer with reward-weighted in-context objectives makes its predictions $M_\theta(\cdot\mid x,D)$ approximate the same Gibbs posteriors $q_{\mathrm{Gibbs}}(\cdot\mid x,D)$ in an amortized fashion.
\end{itemize}

In summary,

\begin{gather}
\text{(i) Few-shot ICL} \;\approx\;
\text{(unweighted) in-context SFT as a forward-KL/MLE projection onto } P_{\mathrm{Bayes}},\\[4pt]
\text{(ii) Reward-weighted ICL (RW-ICL)} \;\approx\;
\text{reward/advantage-weighted in-context SFT},\\[4pt]
\text{(iii) KL-regularized RLHF/RLVR} \;\simeq\;
\text{reward-weighted SFT on the corresponding Gibbs posterior.}
\end{gather}

Here, ``$\approx$'' is intended in an idealized sense at two levels:
at the distribution level, as forward-KL projections onto the same
posterior (Bayes or Gibbs) under our modeling assumptions; and at the
first-order optimization level, as stochastic-gradient updates of the
same weighted-score form described in Section~3.2. It does not claim
that the corresponding practical algorithms are identical beyond these
aspects.

This equivalence is deliberately confined to \emph{objectives}
and \emph{update forms}. The source and granularity of the learning signal remain genuinely
different across paradigms. Supervised fine-tuning observes dense token-level supervision
from mappings $(x, y_{\le i}) \mapsto y_{i+1}$; KL-regularized RL begins with sparse
sequence- or trajectory-level rewards and must perform explicit credit assignment via
value/advantage estimation before reducing to an advantage-weighted SFT update; and few-shot
ICL reuses dense supervision in pretraining to learn a meta-learner $\theta$, but performs
task-specific adaptation at test time in the forward activations rather than via additional
gradients. Our framework reconciles these methods at the level of their posterior targets and
first-order dynamics, while leaving room for these process-level differences to drive distinct
sample-efficiency and stability trade-offs in practice.

All of these procedures are different faces of the same underlying mechanism:
a (possibly generalized) Bayesian update of a reference model that produces a target posterior
(either the true Bayes posterior $P_{\mathrm{Bayes}}$ or a designed Gibbs posterior $q_{\mathrm{Gibbs}}$),
followed by a forward-KL (MLE) projection onto the parameterized policy family.
Meta-learning / first-order interpretations (ICL $\approx$ one-step SFT / AWSFT) can then be seen
as algorithmic realizations of these projection steps and are best viewed as complementary
to the Bayesian backbone developed here.

From an information-design perspective, vanilla SFT passively assumes that the data source already supplies the optimal information structure (the true posterior $q_{\mathrm{true}}$). RLHF/RLVR first designs a target posterior $q_{\mathrm{Gibbs}}$ via the choice of reward $R$ and temperature $\beta$, and then performs the same forward-KL/MLE projection as in SFT. In this sense, RLHF is SFT on a designed information structure: the optimization remains maximum likelihood, but the posterior being fit has been engineered to encode the desired incentives.

The experiments of the preceding section corroborate this picture at the
mechanism level: operators that share their learning-signal granularity
coincide almost exactly when support is good and diverge when it degrades;
cold-start failure appears as a support phenomenon rather than an optimizer
failure; and reward-weighted projection performs on par with its unweighted
counterpart under a matched budget.

\bibliographystyle{plainnat}
\bibliography{refs}

\section*{Appendix A: Bayesian ICL Results Used in the Main Text}

This appendix summarizes the key theoretical results from recent work on Bayesian views of
in-context learning that we rely on in Part~I: \citet{WakayamaBayesICL} and \citet{ReuterBayesTransformer}.
We only state what is needed for our arguments; full technical details and proofs are in the
original papers.

\subsection*{A.1 Risk decomposition and non-asymptotic bounds (Wakayama \& Suzuki)}

\citet{WakayamaBayesICL} formalize a hierarchical prompt-generating process over tasks and
show that the in-context learning (ICL) risk admits an exact orthogonal decomposition under
squared loss. Let $M$ be any in-context predictor (not necessarily a Transformer), and let
$R(M)$ denote its expected ICL risk under their meta-distribution. Then
\begin{equation}
  R(M)
  =
  R_{\mathrm{BG}}(M)
  +
  R_{\mathrm{PV}},
\end{equation}
where:
\begin{itemize}
  \item $R_{\mathrm{BG}}(M)$ is the Bayes Gap: an excess-risk term measuring how far $M$
        is from the Bayes-optimal in-context predictor $M_{\mathrm{Bayes}}$;
  \item $R_{\mathrm{PV}}$ is the Posterior Variance: an irreducible term depending only on
        the task distribution and context length, not on $M$.
\end{itemize}

In particular, $M_{\mathrm{Bayes}}$ is the unique minimizer of $R(M)$, and $R_{\mathrm{PV}}$ captures the
intrinsic task uncertainty. This is exactly the content we use in Theorem~1 in Part~I: the
Bayes posterior predictive is the normative target for in-context prediction, and any in-context
learning algorithm is judged by its Bayes Gap $R_{\mathrm{BG}}(M)$.

For a class of uniform-attention Transformers with feature dimension $m$, context length $p$
and number of pretraining prompts $N$, \citet{WakayamaBayesICL} further provide a
non-asymptotic upper bound of the form
\begin{equation}
  \mathbb{E}\,R_{\mathrm{BG}}(M_{\hat\theta})
  \;\lesssim\;
  m^{-\frac{2\alpha}{d_{\mathrm{eff}}}}
  +
  \widetilde{O}\!\left(\frac{m}{pN}+\frac{1}{N}\right),
\end{equation}
where $\alpha\in(0,1]$ is a H\"older exponent and $d_{\mathrm{eff}}$ is an effective dimension of the
data manifold. The first term is an approximation error controlled by $m$, and the second term
is a pretraining generalization error depending jointly on $p$ and $N$.

They also analyze mixtures of task types and input-distribution shift, showing that:
\begin{itemize}
  \item In mixed-task settings, the posterior over the task index concentrates quickly as the
        context grows, so the Bayes-optimal meta-algorithm rapidly specializes to the true
        task family.
  \item Under input-distribution shift between pretraining and inference, only the Bayes Gap
        increases, with an out-of-distribution penalty proportional to a Wasserstein distance
        between input distributions, while the posterior variance term is intrinsic to the target
        domain.
\end{itemize}

We do not need these bounds explicitly, but they support our assumption in Sec.~1.3 that a
sufficiently expressive Transformer, trained on many tasks, can approximate $M_{\mathrm{Bayes}}$
up to a small Bayes Gap in the regimes of interest.

\subsection*{A.2 Transformers as amortized Bayesian inference (Reuter et al.)}

\citet{ReuterBayesTransformer} study whether in-context Transformers can approximate full Bayesian
posterior inference over latent variables.

They construct several synthetic scenarios (generalized linear models, Gaussian mixture models,
etc.) where the ground-truth posterior over latents is available via high-quality MCMC, and then:
\begin{itemize}
  \item Train an in-context learner $M_\theta$ that maps each dataset $D$ to a predictive
        distribution (or posterior sampler) over latent variables and responses.
  \item Compare the distribution of samples from $M_\theta$ with those from gold-standard
        Bayesian methods (e.g., HMC, SGLD, variational approximations) using metrics such as
        C2ST, MMD, Wasserstein distance, and RMSE.
  \item Investigate robustness under distribution shift between training and test datasets,
        showing that the in-context learner often remains close to the Bayesian baseline in
        out-of-distribution regimes.
\end{itemize}

They find that suitably trained Transformers can act as strong amortized Bayesian
inference engines in context, often matching or approaching fully Bayesian baselines on both
synthetic and real-world datasets, and sometimes outperforming classical approximate methods
in certain distributional metrics.

This provides empirical support for the assumption used in Sec.~1.3 that a sufficiently expressive
Transformer, trained on many tasks, can approximate the Bayes posterior predictive
$P_{\mathrm{Bayes}}(y\mid x,D_k)$ via a single forward pass $M_\theta(y\mid x,D_k)$.

\subsection*{A.3 Relation to the main text (and to meta-learning views)}

In the main body of this note:
\begin{itemize}
  \item Part~I uses \citet{WakayamaBayesICL} to justify that the Bayes posterior predictive is
        the canonical target for in-context prediction, and that ICL risk decomposes into a
        Bayes Gap plus an irreducible posterior-variance term.
  \item The amortized-inference perspective in Sec.~1.3 and the statement that a single
        Transformer forward pass can approximate a dataset-level Bayesian posterior are
        motivated by \citet{ReuterBayesTransformer}.
\end{itemize}

We do not reproduce the gradient-based/meta-learning derivations in this
appendix. Instead, we refer to \citet{AkyurekICLLinear}, \citet{DaiICLMetaOpt}, and \citet{LiTransAsAlgo} for rigorous
analyses showing that, in linear and linearized regimes, Transformers can implement one-step
gradient updates in their forward passes and that attention has an algebraic dual representation
as accumulated gradient corrections. In our view, those results are best interpreted as
algorithmic realizations of the Bayesian predictor described in Parts~I–III, rather than as
the primary conceptual foundation.

\subsection*{A.4 Task--prior representation under within-task exchangeability}

In Part I (Section~1.2), we adopted the hierarchical task--prior model
\[
  f \sim P(f), \qquad (x, y) \mid f \sim p_f(x, y)
\]
to formalize few-shot in-context learning as Bayes posterior prediction over tasks.
In this appendix, we justify that such a hierarchical representation is a canonical consequence
of a much weaker and more intuitive assumption on within-task data: \emph{exchangeability}.
Throughout, exchangeability is an assumption on the \emph{data-generating process} within
a task episode, not on any particular neural architecture used to implement in-context
learning.

We first recall the notion of exchangeability and then invoke a classical representation theorem
of de Finetti--Hewitt--Savage to obtain the hierarchical model used in the main text.

\paragraph{A.4.1 Within-task exchangeability.}

Let $\mathcal{X}$ be the input space and $\mathcal{Y}$ the output space (e.g., labels or next tokens).
Define the sample space $\mathcal{Z} := \mathcal{X} \times \mathcal{Y}$.
We consider sequences
\[
  Z_i = (X_i, Y_i) \in \mathcal{Z}, \qquad i = 1, 2, \dots
\]
generated under a single ``task episode''.

\textbf{Definition A.4.1 (Finite and infinite exchangeability).}
A finite sequence $Z_1, \dots, Z_n$ is \emph{exchangeable} if for every permutation
$\pi$ of $\{1, \dots, n\}$,
\[
  (Z_1, \dots, Z_n) \stackrel{d}{=} (Z_{\pi(1)}, \dots, Z_{\pi(n)}),
\]
i.e., its joint distribution is invariant under any reordering of the indices.

An infinite sequence $(Z_i)_{i \ge 1}$ is \emph{exchangeable} if for every $n \ge 1$,
the finite prefix $(Z_1, \dots, Z_n)$ is exchangeable.

Intuitively, for an exchangeable sequence, only the multiset
$\{Z_1, \dots, Z_n\}$ matters, not the order.
This is a standard formalization of ``order-invariant'' data within a task.
Note that i.i.d.\ sequences are a special case of exchangeable sequences,
but exchangeability is strictly weaker.

In our setting, we model each task episode as inducing an exchangeable sequence of sample pairs:

\textbf{Assumption A.4.2 (Within-task exchangeability).}
For each task episode, the (conceptual) infinite sequence of observations
$(Z_i)_{i \ge 1} = (X_i, Y_i)_{i \ge 1}$ taking values in
$\mathcal{Z} = \mathcal{X} \times \mathcal{Y}$ is exchangeable.

In practice, we only ever observe a finite dataset
$D_k = \{(x_i, y_i)\}_{i=1}^k$, but we assume it arises as a finite prefix
of such an exchangeable sequence.
This is the usual idealization used in probability theory:
we posit that, if we were to continue sampling from the same task,
the joint law of the extended sequence would remain exchangeable.

\paragraph{A.4.2 A de Finetti--Hewitt--Savage representation.}

The sample space $\mathcal{Z} = \mathcal{X} \times \mathcal{Y}$ is a standard Borel space in typical
supervised learning settings (e.g., $\mathcal{X}, \mathcal{Y}$ finite or subsets of $\mathbb{R}^d$),
so we can invoke the general form of de Finetti's representation theorem as extended by
Hewitt and Savage.

We state an informal version sufficient for our purposes.

\textbf{Theorem A.4.3 (de Finetti--Hewitt--Savage representation; informal).}
Let $(Z_i)_{i \ge 1}$ be an infinite exchangeable sequence of random variables
taking values in a standard Borel space $\mathcal{Z}$.
Then there exists a random probability measure $\mu$ on $\mathcal{Z}$
(a random element of $\mathcal{P}(\mathcal{Z})$) such that,
conditional on $\mu$, the sequence is i.i.d.\ with common distribution $\mu$:
\[
  (Z_i)_{i \ge 1} \mid \mu \;\text{i.i.d.} \sim \mu, \qquad \mu \sim \Pi,
\]
for some probability law $\Pi$ on $\mathcal{P}(\mathcal{Z})$.
Moreover, the mixing law $\Pi$ is unique.

Equivalently, the joint law of $(Z_i)_{i \ge 1}$ can be written as
a \emph{mixture of i.i.d.\ sequences}.

The original versions of this theorem were proved by de Finetti for Bernoulli sequences and
generalized by Hewitt, Savage, and many others to arbitrary standard Borel spaces;
see, e.g., standard modern references in probability theory for precise statements and proofs.

Under Assumption~A.4.2, the sequence $(Z_i)$ generated by a given task episode
satisfies the hypotheses of Theorem~A.4.3, and hence admits such a mixture-of-i.i.d.\ representation.

\paragraph{A.4.3 From a random measure to a task prior.}

We now reinterpret the random measure $\mu$ from Theorem~A.4.3 as a \emph{task},
and its mixing distribution $\Pi$ as a \emph{task prior} $P(f)$.

Let $\mathcal{F} := \mathcal{P}(\mathcal{Z})$ denote the space of all probability measures
on $\mathcal{Z}$, equipped with the $\sigma$-algebra generated by the usual weak topology.
For each realization $\mu \in \mathcal{F}$, define the joint law on $\mathcal{Z}$ by
\[
  p_\mu(z) := \mu(\{z\}), \qquad z \in \mathcal{Z},
\]
or more generally $p_\mu(z)$ as the Radon--Nikodym density of $\mu$ with respect
to a suitable base measure if $\mathcal{Z}$ is continuous.

We introduce the following identification:
\begin{itemize}
  \item \textbf{Tasks.} Each realization of $\mu$ is interpreted as a task $f \in \mathcal{F}$,
        with joint law $p_f(x, y)$ over $\mathcal{X} \times \mathcal{Y}$;
  \item \textbf{Task prior.} The mixing law $\Pi$ over $\mu$ is renamed as a prior $P(f)$ over tasks.
\end{itemize}

Under this identification, Theorem~A.4.3 can be restated as:

\begin{quote}
There exists a random task $f \sim P(f)$ and a conditional distribution
$(X_i, Y_i) \mid f \sim p_f(x, y)$ such that, for each task episode, the joint law
of $(Z_i)_{i \ge 1} = (X_i, Y_i)_{i \ge 1}$ is the same as that of an i.i.d.\ sequence
drawn from $p_f$.
\end{quote}

Formally, we obtain the hierarchical representation:
\[
  f \sim P(f), \qquad (X_i, Y_i)_{i \ge 1} \mid f \ \text{i.i.d.} \sim p_f(x, y).
\]

This is exactly the \emph{task--prior generative model} used in Section~1.2
and the main Bayesian ICL analysis.

For supervised tasks, it is often convenient to factor $p_f(x, y)$ as
\[
  p_f(x, y) = p_f(x)\, p_f(y \mid x),
\]
and to interpret $f$ primarily via its conditional component $p_f(y \mid x)$.
In that case, we may succinctly write $f : \mathcal{X} \to \Delta(\mathcal{Y})$,
with $P(f)$ a prior over such conditional task functions, as in the main text.

We summarize this as a proposition:

\textbf{Proposition A.4.4 (Task--prior representation under exchangeability).}
Suppose that, for each task episode, the within-task observation sequence
$(X_i, Y_i)_{i \ge 1}$ taking values in $\mathcal{X} \times \mathcal{Y}$ is infinite
and exchangeable (Assumption~A.4.2).
Then there exists a task space $\mathcal{F}$ and a prior $P(f)$ over $\mathcal{F}$
such that the joint distribution over data induced by a task episode admits
the hierarchical representation
\[
  f \sim P(f), \qquad (X_i, Y_i) \mid f \ \text{i.i.d.} \sim p_f(x, y).
\]
In particular, the existence of a task prior $P(f)$ is not an additional modeling assumption
beyond within-task exchangeability; it is a canonical hierarchical reparameterization
of the same joint law guaranteed by the de Finetti--Hewitt--Savage theorem.

\emph{Proof sketch.}
Apply Theorem~A.4.3 to the exchangeable sequence
$(Z_i)_{i \ge 1} = (X_i, Y_i)_{i \ge 1}$ on $\mathcal{Z} = \mathcal{X} \times \mathcal{Y}$
to obtain a random measure $\mu$ on $\mathcal{Z}$ and a mixing law $\Pi$
such that $(Z_i) \mid \mu$ are i.i.d.\ with common distribution $\mu$, and $\mu \sim \Pi$.
Define $\mathcal{F} := \mathcal{P}(\mathcal{Z})$, identify each $\mu \in \mathcal{F}$
with a task $f$, and set $P(f) := \Pi$.
Let $p_f(x, y)$ denote the joint law induced by $f$ on $\mathcal{X} \times \mathcal{Y}$.
Then
\[
  f \sim P(f), \qquad (X_i, Y_i) \mid f \ \text{i.i.d.} \sim p_f(x, y)
\]
is exactly the hierarchical generative process whose marginal law over
$(X_i, Y_i)_{i \ge 1}$ coincides with the original exchangeable sequence.
\hfill$\square$

\paragraph{A.4.4 Remarks and scope.}

\textbf{(1) What is the real modeling assumption?}
Proposition~A.4.4 shows that the only substantive structural assumption we make at the task level
is within-task exchangeability (Assumption~A.4.2).
Given this, the existence of a hierarchical task prior $P(f)$ is a representation theorem,
not an extra ad hoc assumption.

\textbf{(2) Finite sequences and approximate exchangeability.}
In practice, we only observe finite datasets $D_k = \{(x_i, y_i)\}_{i=1}^k$.
The infinite-sequence formulation assumes that these finite datasets can be extended
to an infinite exchangeable sequence.
There also exist ``finite de Finetti'' theorems which provide approximate mixture
representations for finitely exchangeable sequences; we do not rely on their technical details here,
but conceptually they support the view that approximately exchangeable finite datasets
are well-approximated by such hierarchical models.

\textbf{(3) Connection to Wakayama \& Suzuki's hierarchical model.}
Wakayama and Suzuki~[14] start directly from a hierarchical task prior $f \sim P(f)$
and derive non-asymptotic generalization bounds and a Bayes-optimal ICL risk decomposition.
Our use of de Finetti--type representations clarifies that this hierarchical structure
is not arbitrary: it is the canonical form of the joint distribution under within-task exchangeability.
This provides a principled foundation for adopting their model in Part~I.

\textbf{(4) Empirical plausibility.}
Whether within-task exchangeability is a good approximation for real pretraining corpora
is ultimately an empirical question.
In the main text, we view the task--prior model as an \emph{effective theory} of the latent task
structure in natural data and discuss experimental designs (e.g., shuffling document structure,
mixing tasks within prompts) that can probe the degree to which such a model is appropriate.

\textbf{(5) Order effects in practical in-context learners.}
Assumption~A.4.2 and Proposition~A.4.4 concern the joint law of within-task data
$(X_i, Y_i)_{i\ge 1}$ under an exchangeable data-generating process. Real transformer
models process ordered prompts with positional embeddings, and the order of demonstrations
can affect predictions (e.g., via position-dependent attention weights). As discussed in
Section~1.2, we treat such order effects as implementation-level deviations from the
exchangeable Bayes benchmark characterized here, rather than violations of the underlying
exchangeability assumption.

\end{document}